\documentclass[10pt,final,journal,letterpaper,oneside,twocolumn]{IEEEtran}

\usepackage{cite}
\usepackage[pdftex]{graphicx}
\usepackage[cmex10]{amsmath}
\usepackage{amssymb}
\usepackage{amsfonts}
\usepackage{algorithm}
\usepackage{algorithmic}
\usepackage[caption=false,font=footnotesize]{subfig}
\usepackage{color}
\usepackage{array}
\usepackage{mdwmath}
\usepackage{mdwtab}
\usepackage{eqparbox}
\usepackage{lineno}

\begin{document}

%\pagewiselinenumbers

%\title{Cognitive Internet of Things: \\ Only Connected is Not Enough}

\title{Cognitive Internet of Things: \\ A New Paradigm beyond Connection}

\author{\IEEEauthorblockN{Qihui~Wu,~\emph{Senior Member, IEEE}, Guoru Ding,~\emph{Student Member, IEEE}, Yuhua Xu,~\emph{Student Member, IEEE},\\
Shuo Feng, Zhiyong Du, Jinlong Wang,~\emph{Senior Member, IEEE}, and Keping Long,~\emph{Senior Member, IEEE}}

\date{}

\thanks{This work has been accepted for publication by \emph{IEEE Journal of Internet of Things}. Personal use of the material in this work is permitted. Permission from IEEE must be obtained for all other uses, including reprinting/republishing this material for advertising or promotional purposes, collecting new collected works for resale or redistribution to servers or lists, or reuse of any copyrighted component of this work in other works.}
\thanks{This work is supported by the National Natural Science Foundation of China (Grant No. 61172062, 61301160) and in part by Jiangsu Province Natural Science Foundation (Grant No. BK2011116).}
\thanks{Q. Wu, G. Ding, Y. Xu, S. Feng, Z. Du, and J. Wang are with the College of Communications Engineering, PLA University of Science and Technology, Nanjing 210007, China (email: wqhtxdk@aliyun.com; dingguoru@gmail.com; yuhuaenator@gmail.com; duzhiyong2006@163.com; fengshuo1010@163.com; wjl543@sina.com). G. Ding is the corresponding author.}
\thanks{K. Long is with Institute of Advanced Network Technologies and New Services (ANTS) and Beijing Engineering and Technology Center for Convergence Networks and Ubiquitous Services, University of Science and Technology Beijing (USTB), No. 30, Xueyuan Road, Haidian District, Beijing, China 100083 (e-mail: longkeping@ustb.edu.cn).}}

\maketitle

\begin{abstract}
Current research on Internet of Things (IoT) mainly focuses on how to enable general objects to see, hear, and smell the physical world for themselves, and make them connected to share the observations. In this paper, we argue that only connected is not enough, beyond that, general objects should have the capability to \emph{learn}, \emph{think}, and \emph{understand} both physical and social worlds by themselves. This practical need impels us to develop a new paradigm, named Cognitive Internet of Things (CIoT), to empower the current IoT with a `brain' for high-level intelligence. Specifically, we first present a comprehensive definition for CIoT, primarily inspired by the effectiveness of human cognition. Then, we propose an operational framework of CIoT, which mainly characterizes the interactions among five fundamental cognitive tasks: perception-action cycle, massive data analytics, semantic derivation and knowledge discovery, intelligent decision-making, and on-demand service provisioning. Furthermore, we provide a systematic tutorial on key enabling techniques involved in the cognitive tasks. In addition, we also discuss the design of proper performance metrics on evaluating the enabling techniques. Last but not least, we present the research challenges and open issues ahead. Building on the present work and potentially fruitful future studies, CIoT has the capability to bridge the physical world (with objects, resources, etc.) and the social world (with human demand, social behavior, etc.), and enhance smart resource allocation, automatic network operation, and intelligent service provisioning.
\end{abstract}

\IEEEpeerreviewmaketitle


\section{Introduction}

\subsection{Background and Motivation}
The \emph{Internet of Things (IoT)}, firstly coined by Kevin Ashton as the title of a presentation in 1999~\cite{Ashton-1999}, is a technological revolution that is bringing us into a new ubiquitous connectivity, computing, and communication era. The development of IoT depends on dynamic technical innovations in a number of fields, from wireless sensors to nanotechnology~\cite{ITU-2005}. For these ground-breaking innovations to grow from ideas to specific products or applications, in the past decade, we have witnessed worldwide efforts from academic community, service providers, network operators, and standard development organizations, etc (see, e.g., the recent comprehensive surveys in~\cite{ASurvey,Context-Survey,Standardized-survey}). Technically, most of the attention has been focused on aspects such as communication, computing, and connectivity, etc, which are indeed very important topics. However, we argue that without comprehensive cognitive capability, IoT is just like an awkward stegosaurus: all brawn, no brains. To fulfill its potential and deal with growing challenges, we must take the cognitive capability into consideration and empower IoT with high-level intelligence. Specifically, in this paper, we develop an enhanced IoT paradigm, i.e., \emph{Brain-Empowered Internet of Things} or \emph{Cognitive Internet of Things (CIoT)}, and investigate the involved key enabling techniques.

Before gonging deep into the new concept CIoT and its enabling techniques, let's first share two interesting application scenarios that will probably come into our daily life in future:

\emph{Application scenario 1:} Let's imagine that it's Friday, after five days' hard work, I'd like to relax myself and watch a TV Soap Opera tonight. When time goes to the midnight, I become more and more sleepy and finally fall asleep on my sofa. Generally, I will wake up late on Saturday and feel very tired since I do not sleep well with the TV noise, the uncomfortable sofa and the fluctuating temperature all night long. Consequently, I have a dream that one day the TV, the sofa, and the air conditioner in my room could individually or cooperatively sense my movement, gesture, and/or voice, based on which they analyze my state (e.g., `sleepy' or `not sleepy'), and make corresponding decisions by themselves to comfort me, e.g., if I am in the state of `sleepy', the TV itself gradually lowers or even turns off the voice, the sofa slowly changes itself to a bed, and the air conditioner dynamically adjusts the temperature suitable for sleep.

%---------------------------------------------------------------------------------------------------------------------------------------------------
\begin{figure*}[!t]
\centerline{
\subfloat[$\text{Application scenario 1}$.]{\includegraphics[width=2.5in]{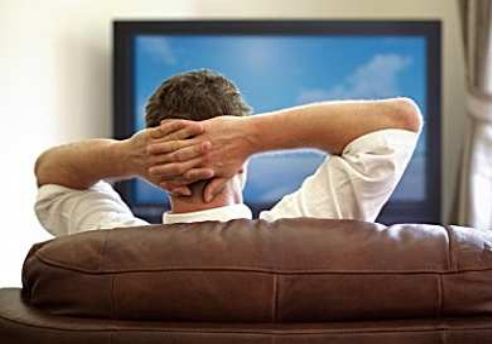}
\label{Fig:Channel}}
\hfil
\subfloat[$\text{Application scenario 2}$.]{\includegraphics[width=2.5in]{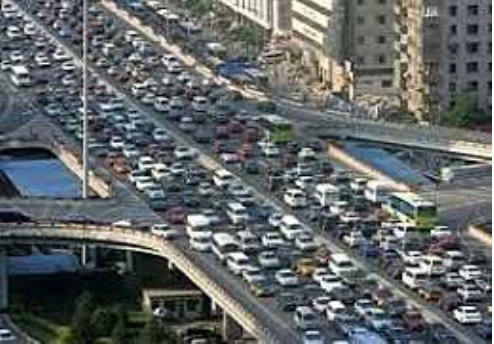}
\label{Fig1:Abnormal-Strength}}
}
\caption{Motivational and illustrative application scenarios.}
\label{fig_comparison}
\end{figure*}
%---------------------------------------------------------------------------------------------------------------------------------------------------

\emph{Application scenario 2:} Living in a modern city, traffic jams harass many of us. With potential traffic jams into consideration, every time when the source and the destination is clear, it is generally not easy for a driver to decide what the quickest route should be, especially when the driver is fresh to the city. Among many others, the following scheme may be welcome and useful for drivers: Suppose that there are a city of crowdsourcers, such as pre-deployed cameras, vehicles, drivers, and/or passengers, intermittently observe the traffic flow nearby and contribute their observations to a data center. The data center effectively fuses the crowdsourced observations to generate real-time traffic situation map and/or statistical traffic database. Then, every time when a driver tells his/her car the destination, the car will automatically query the data center, deeply analyze the accessed traffic situation information from the data center and meanwhile other cars/drivers' potential decisions, and intelligently selects the quickest route or a few top quickest routes for its driver.

Just like the aforementioned examples, you may be familiar with a lot of other blueprints of ``intelligent life" and look forward to that the dreams could become true soon. But things are not that simple as they look like. Just as a sportscar without engine is only a gorgeous waste, the IoT without a `brain' is not enough to bring the expected convenient and comfortable life to us. These observations motivate us to develop the new paradigm \emph{Cognitive Internet of Things (CIoT)}.

But, first and foremost, what do we mean by Cognitive Internet of Things? Before responding to this question, it is in order that we first address the meaning of the related term ``cognition." Referring from the well-known books~\cite{Cognition,SHaykin-CDS,Intelligence,Intelligence2}, it is more appropriate to refer to ``cognition" as an ``integrative field" rather than a ``discipline" since the study on ``cognition" integrates many fields that are rooted in neuroscience, cognitive science, computer science, mathematics, physics, and engineering, etc. Specifically, in this paper, the authors take the operational process of human brain as the reference framework for cognition~\cite{SHaykin-CDS}, and offer the following definition for cognitive internet of things:

\emph{Cognitive Internet of Things (CIoT) is a new network paradigm, where (physical/virtual) things or objects are interconnected and behave as agents, with minimum human intervention, the things interact with each other following a context-aware perception-action cycle, use the methodology of understanding-by-building to learn from both the physical environment and social networks, store the learned semantic and/or knowledge in kinds of databases, and adapt themselves to changes or uncertainties via resource-efficient decision-making mechanisms, with two primary objectives in mind:
\begin{itemize}
  \item bridging the physical world (with objects, resources, etc) and the social world (with human demand, social behavior, etc), together with themselves to form an intelligent physical-cyber-social (iPCS) system;
  %\item providing ubiquitous connectivity and computing, and highly reliable communication whenever and wherever needed;
  \item enabling smart resource allocation, automatic network operation, and intelligent service provisioning.
\end{itemize}}

\subsection{Historical Notes}
The history of \emph{Internet} goes back to the development of communication between two computers through a computer network in the late 1960s~\cite{network}. Since then, the evolution of the Internet has passed three main phases: Internet of Computers, Internet of People (mainly via social networking), and Internet of Things (including computers, people and any other physical/virtual objects).

As mentioned above, the term `\emph{Internet of Things}' was firstly coined by Kevin Ashton in 1999~\cite{Ashton-1999}. Then, in 2001 the MIT Auto-ID center presented their Internet of Things (IoT) vision~\cite{MIT-Auto-ID}. Later, in 2005 IoT was formally introduced as the theme of the seventh in the series of International Telecommunication Union (ITU) Internet reports~\cite{ITU-2005}. In 2008, the first international conference on the internet of things was held in Zurich~\cite{IoT-2008}. In 2009, China government advocated the idea of ``Sensing China" and Wuxi city became one of the leading centers of IoT-related research and industry in China~\cite{Wuxi-IoT}. At the same year, IoT European Research Cluster (IERC) presented a document of IoT strategic research roadmap on future research and development until 2015 and beyond 2020~\cite{roadmap}, and one year later, published a comprehensive document on the vision and challenges for realizing the IoT~\cite{CERP-IoT}. In the past couple of years, the IoT has gained significantly increasing attention from academia as well as industry, comprehensive surveys can be found in~\cite{ASurvey,Context-Survey,Standardized-survey}. Briefly, so far IoT is a very broad paradigm and many visions (e.g., ``Internet oriented visions," ``Things oriented visions," and ``Semantic oriented visions"~\cite{ASurvey}) coexist.

Unlike (conventional) IoT, the research on \emph{Cognitive Internet of Things (CIoT)} is very limited. In~\cite{cognitive-management-framework}, a cognitive management framework is presented to empower the IoT to better support sustainable smart city development, where cognition mainly refers to the autonomic selection of the most relevant objects for the given application. In~\cite{CIoT-Henan}, CIoT is viewed as the current IoT integrated with cognitive and cooperative mechanisms to promote performance and achieve intelligence, where the cognitive process is made up of a three-layer cognitive ring. Alternatively, in this paper we coin the CIoT by integrating the operational process of human cognition into the system design, and we also provide systematic discussions on the key enabling techniques for the fundamental cognitive tasks involved in the research and development of CIoT.

Another related topic is~\emph{Cognitive Radio Networks (CRN)}, which was firstly proposed by Joseph Mitola III in 1999~\cite{Mitola-1999} and recoined by Simon Haykin in 2005 from a signal processing perspective~\cite{Haykin_2005}, and since then the research on CRN has been one of the hottest topics in the field of wireless communications (see, e.g.,~\cite{IFA-2006,CRN-SPM,CRN-TVT,Gao2011TVT,TWC2012}). One common point of CIoT and CRN is that both of them benefit from the recent advances in cognitive science~\cite{Cognition,SHaykin-CDS,Intelligence,Intelligence2}. The differences between CIoT and CRN are much more than the common. CRN is well-known as a promising paradigm to improve the utilization of radio electromagnetic spectrum, by allowing unlicensed radios to opportunistically access the idle spectrum licensed to the primary radios~\cite{Mitola-1999,IFA-2006,Haykin_2005,CRN-SPM,CRN-TVT,Gao2011TVT,TWC2012,LRP_Cmag}. CRN is in essence a radio system with the objective to improve wireless network throughput. However, CIoT generally consists of (massive) heterogeneous general objects, not just radios, with various objectives for different applications. Moreover, the technical research on CIoT should not focus on specific applications, instead, it should be general enough to support as many applications as possible, and consequently face much more unique challenges, which will be discussed in detail in the following sections.

%%------------------------------------------------------------------------------------------
\begin{figure*}[!t]
\centering
\includegraphics[width=0.8\linewidth]{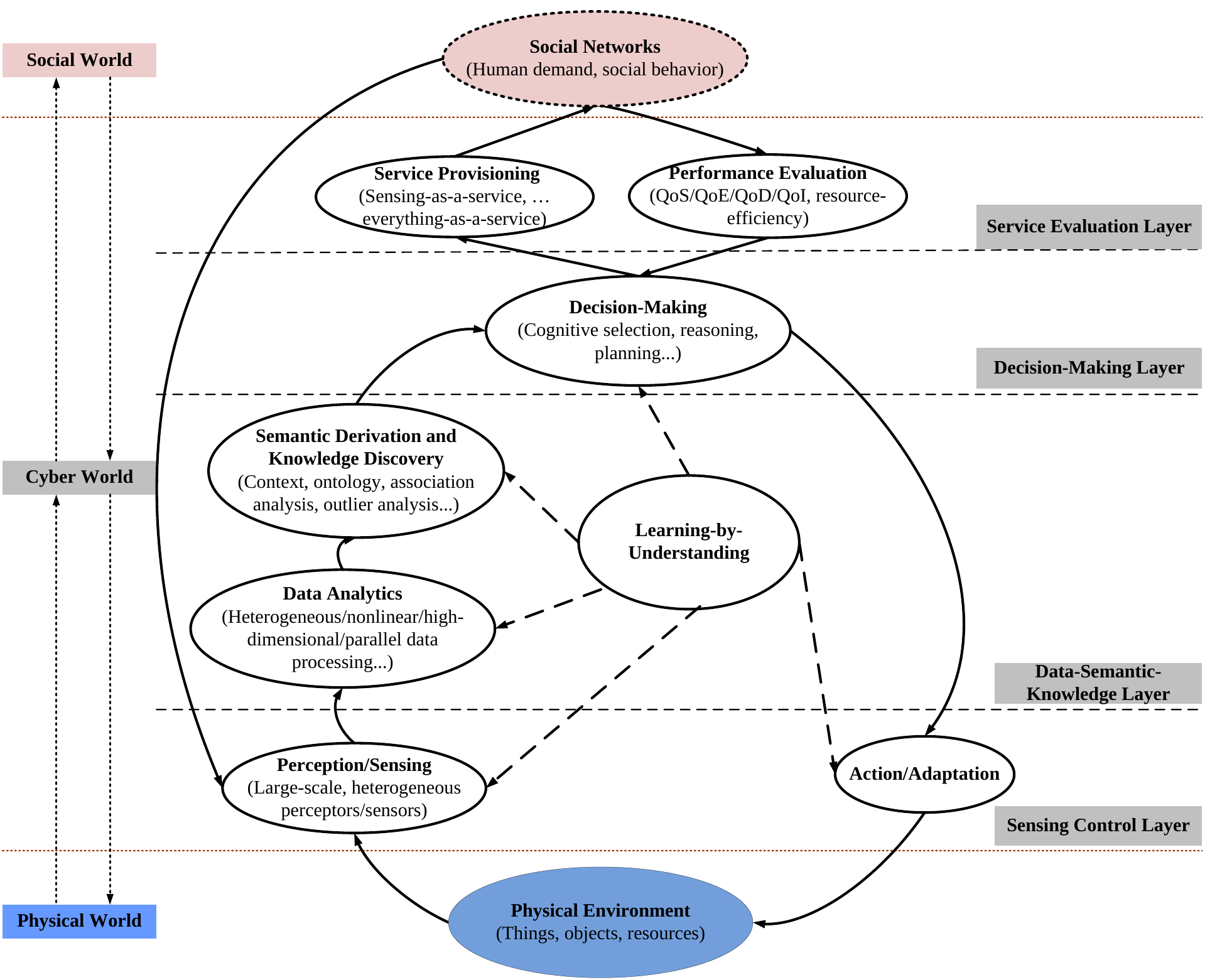}
\caption{Framework of Cognitive Internet of Things (CIoT).}
\label{Function}
\end{figure*}
%%-----------------------------------------------------------------------------------------

\subsection{Purpose of this Paper}
The original motivation of the concept `Internet of Things' was explained by Kevin Ashton as follows~\cite{Ashton-1999}:

``Today computers-and, therefore, the Internet-are almost wholly dependent on human beings for information ... The problem is, people have limited time, attention and accuracy ... We need to empower computers with their own means of gathering information, so they can see, hear and smell the world for themselves..."

The primary purpose of this paper is to build on Kevin Ashton's visionary insights and enhance them by empower general objects to \emph{learn}, \emph{think}, and \emph{understand} physical and social worlds by themselves, by effectively integrating the operational process of human cognition into the design of IoT and presenting detailed expositions of cognitive processing techniques that lie at the heart of Cognitive Internet of Things.

\subsection{Organization and Potential Applications}
The reminder of this paper is organized as follows. Section II presents an overview of CIoT. Section III-V sequentially address the key enabling techniques for the fundamental cognitive tasks. Section VI provides discussion on the design of performance evaluation metrics for CIoT. Section VII presents the research challenges and open issues and Section VIII concludes the paper.

The work in this paper can be applied to many practical applications, e.g., the two application scenarios (i.e., smart TV and intelligent transportation) described in Section I-A. Taking the second application scenario as an example, the framework of CIoT developed in this Section can be applied to build the architecture of an intelligent transportation system, and the enabling techniques introduced in Section III-V can be embedded to CIoT companies' products, such as the software or Apps.

\section{Cognitive Internet of Things: An Overview}

\subsection{From Internet of Things to Cognitive Internet of Things}
Currently, one of the most distinguished characteristics of Internet of Things is that: with the increasing inter-connectivity among general things or objects, a number of interesting services or applications are emerging. However, so far many of the existing Internet of Things applications are still dependent highly on human beings for cognition processing. This observation serves as one of the primary motivations of this paper to introduce `Cognitive Internet of Things', where general objects behave as agents, and interact with physical environment and/or social networks, with \emph{minimum human intervention}. Briefly, Cognitive Internet of Things enhances the current Internet of Things by mainly integrating the human cognition process into the system design. The advantages are multi-fold, e.g., saving people's time and effort, increasing resource efficiency, and enhancing service provisioning, to just name a few.

\subsection{Framework of CIoT and Fundamental Cognitive Tasks}
Fig. \ref{Function} presents a framework of CIoT. Generally, CIoT serves as a \emph{transparent bridge} between physical world (with general physical/virtual things, objects, resources, etc.) and social world (with human demand, social behavior, etc.), together with itself form an \emph{intelligent physical-cyber-social (iPCS)} system. From a \emph{bottom-up} view, the cognitive process of the iPCS system consists of four major layers:
\begin{itemize}
  \item \emph{Sensing control layer} has direct interfaces with physical environment, in which the perceptors sense the environment by processing the incoming stimuli and feedbacks observations to the upper layer, and the actuators act so as to control the perceptors via the environment.
  \item \emph{Data-semantic-knowledge layer} effectively analyzes the sensing data to form useful semantic and knowledge.
  \item \emph{Decision-making layer} uses the semantic and knowledge abstracted from the lower layer to enable multiple or even massive interactive agents to reason, plan and select the most suitable action, with dual functions to support services for human/social networks and stimulate action/adapation to physical environment.
  \item \emph{Service evaluation layer} shares important interfaces with social networks, in which on-demand service provisioning is provided to social networks, and novel performance metrics are designed to evaluate the provisioned services and feedback the evaluation result to the cognition process.
\end{itemize}

With a synthetic methodology \emph{learning-by-understanding} located at the heart, the framework of CIoT includes five fundamental cognitive tasks, sequentially, \emph{Perception-action cycle}, \emph{Massive data analytics}, \emph{Semantic derivation and knowledge discovery}, \emph{Intelligent decision-making}, and \emph{On-demand service provisioning}. Briefly, perception-action cycle is the most primitive cognitive task in CIoT with perception as the input from the physical environment and action as the output to it. On the other hand, on-demand service provisioning directly supports various services (e.g., Infrastructure-as-a-Service (IaaS), Platform-as-a-Service (PaaS), Sensing-as-a-Service (SaaS), and more broadly Everything-as-a-service (EaaS)~\cite{EaaS}) to human/social networks, which has been investigated recently (see, e.g.,~\cite{SOA-2010,Semantic-service-2011,SaaS}). In the following sections, we will focus on the key enabling techniques involved in the other three fundamental cognitive tasks.

\section{Massive Data Analytics in Cognitive Internet of Things}

The future CIoT will be highly populated by large numbers of heterogeneous interconnected embedded devices, which are generating massive data in an explosive fashion. The data we collect may not have any value unless we analyze, interpret, understand, and properly exploit it. Taking the application scenario 2 introduced in Section I-B as an example, the traffic data is collected from massive crowdsourcers, including pre-deployed cameras, vehicles, drivers, and passengers, which are generally noisy, corrupted, heterogeneous, high-dimensional, and nonlinear separable. To exploit the value of the massive data, the development of effective algorithms on massive data analytics is urgently needed.

%%------------------------------------------------------------------------------------------
\begin{figure}[!b]
\centering
\includegraphics[width=\linewidth]{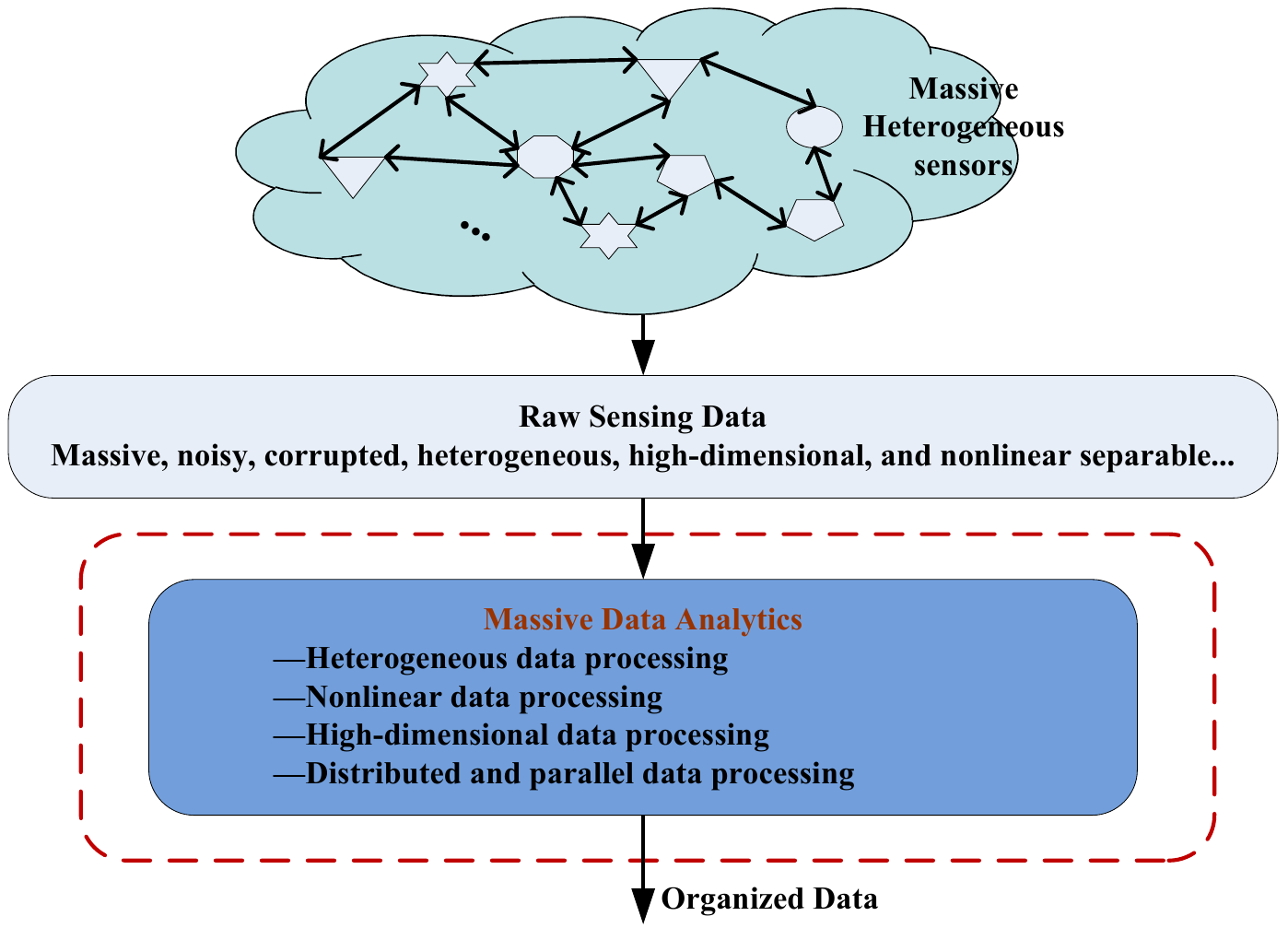}
\caption{The framework of massive data analytics in CIoT.}
\label{Visio-framework-massive}
\end{figure}
%%-----------------------------------------------------------------------------------------

As shown in Fig.~\ref{Visio-framework-massive}, in this section we propose a systematic tutorial on the development of effective algorithms for massive data analytics, which are grouped into four classes: 1) heterogeneous data processing, 2) nonlinear data processing, 3) high-dimensional data processing, and 4) distributed and parallel data processing.

\subsection{Heterogeneous Data Processing}

In practical CIoT applications, the massive data are generally collected from heterogeneous sensors (e.g., cameras, vehicles, drivers, and passengers), which in turn may provide heterogeneous sensing data (e.g., text, video, and voice). Heterogeneous data processing (e.g., fusion, classification) brings unique challenges and also offers several advantages and new possibilities for system improvement.

Mathematically, random variables that characterize the data from heterogeneous sensors may follow disparate probability distributions. Denote $z_n$ as the data from the $n$-th sensor and ${\bf{Z}}:=\{z_n\}_{n=1}^N$ as the heterogeneous data set, the marginals $\{z_n\}_{n=1}^N$ are generally non-identically or heterogeneously distributed. In many CIoT applications, problems are often modeled as multi-sensor data fusion, distribution estimation or distributed detection. In these cases, joint probability density function (pdf) $f({\bf{Z}})$ of the heterogeneous data set ${\bf{Z}}$ is needed to obtain from the marginal pdfs $\{f(z_n)\}_{n=1}^N$.

For mathematical tractability, one often chooses to assume simple models such as the product model or multivariate Gaussian model, which lead to suboptimal solutions~\cite{Correlation}. Here we recommend another approach, based on copula theory, to tackle heterogeneous data processing in CIoT. In copula theory, it is the copulas function that couples multivariate joint distributions to their marginal distribution functions, mainly thanks to the following theorem:

\emph{Sklar' Theorem~\cite{copula}:} Let $F$ be an $N$-dimensional cumulative distribution function (cdf) with continuous marginal cdfs $F_1,F_2,...,F_N$. Then there exists a unique copulas function $C$ such that for all $z_1,z_2,...,z_N$ in $[-\infty,+\infty]$
\begin{align}
\label{eq:Copula}
F(z_1,z_2,...,z_N)=C\big(F_1(z_1),F_2(z_2),...,F_N(z_N)\big).
\end{align}

The joint pdf can now be obtained by taking the $N$-order derivative of (\ref{eq:Copula})
\begin{align}
\label{eq:Copula2}
&f(z_1,z_2,...,z_N)\nonumber\\
&=\frac{\partial^N}{\partial_{z_1}\partial_{z_2}...\partial_{z_N}}C\big(F_1(z_1),F_2(z_2),...,F_N(z_N)\big)\nonumber\\
&=f_p(z_1,z_2,...,z_N)c\big(F_1(z_1),F_2(z_2),...,F_N(z_N)\big),
\end{align}
where $f_p(z_1,z_2,...,z_N)$ denotes the product of the marginal pdfs $\{f(z_n)\}_{n=1}^N$ and $c(\cdot)$ is the copula density weights the product distribution appropriately to incorporate dependence between the random variables. The topic on the design or selection of proper copula functions is well summarized in~\cite{copula2}.

\subsection{Nonlinear Data Processing}

In CIoT applications, such as multi-sensor data fusion, the optimal fusion rule can be derived from the multivariate joint distributions obtained in (\ref{eq:Copula2}). However, it is generally mathematically intractable since the optimal rule generally involves nonlinear operations~\cite{Data-Fusion}. Therefore, linear data processing methods dominate the research and development, mainly for their simplicity. However, linear methods are often oversimplified to deviate the optimality.

In many practical applications, nonlinear data processing significantly outperforms their linear counterparts. Kernel-based learning (KBL) provides an elegant mathematical means to construct powerful nonlinear variants of most well-known statistical linear techniques, which has recently become prevalent in many engineering applications~\cite{SPMag2013}.

Briefly, in KBL theory, data ${\bf{x}}$ in the input space $\mathcal{X}$ is projected onto a higher dimensional feature
space $\mathcal{F}$ via a \emph{nonlinear mapping} $\Phi$ as follows:
\begin{align}
\label{eq:nonlinear-map}
\Phi:~\mathcal{X} ~ \to ~\mathcal{F},~~~~{\bf{x}}~ \mapsto ~ \Phi({\bf{x}}).
\end{align}

For a given problem, one now works with the mapped data $\Phi({\bf{x}}) \in \mathcal{F}$ instead of ${\bf{x}} \in \mathcal{X}$. The data in the input space can be projected onto different feature spaces with different mappings. The diversity of feature spaces provides us more choices to gain better performance. Actually, without knowing the mapping $\Phi$ explicitly, one only needs to replace the inner product operator of a linear technique with an appropriate~\emph{kernel} ${\rm{k}}$ (i.e., a positive semi-definite symmetric function),
\begin{align}
\label{eq:kernel}
{\rm{k}}({\bf{x}}_i,{\bf{x}}_j) := \langle\Phi({\bf{x}}_i), \Phi({\bf{x}}_j)\rangle_{\mathcal{F}}, ~~\forall {\bf{x}}_i,{\bf{x}}_j \in\mathcal{X}.
\end{align}

The most widely used kernels can be divided into two categories: projective kernels (functions of inner product, e.g., polynomial kernels) and radial kernels (functions of distance, e.g., Gaussian kernels)~\cite{SPMag2013}.

\subsection{High-Dimensional Data Processing}
%Data can be collected from different locations and various time instants

In CIoT, massive data always accompanies high-dimensionality. For example, images and videos observed by cameras in many CIoT applications are generally very high-dimensional data, where the dimensionality of each observation is comparable to or even larger than the number of observations. Moreover, in kernel-based learning methods discussed above, the kernel function nonlinearly maps the data in the original space into a higher dimensional feature space, which transforms virtually every dataset to a high-dimensional one.

Mathematically, we can represent the massive data in a compact matrix form. Many practical applications have experimentally demonstrated the intrinsic low-rank property of the high-dimensional data matrix, such as the traffic matrix in large scale networks~\cite{Tracking-Network} and image frame matrix in video surveillance~\cite{RPCA}, which is mainly due to common temporal patterns across columns or rows, and periodic behavior across time, etc.

Low-rank matrix plays a central role in large-scale data analysis and dimensionality reduction. In the following, we provide a brief tutorial on using low-rank matrix recovery and/or completion\footnote{Matrix completion aims to recover the missing entries of a matrix, given limited number of known entries, while matrix recovery aims to recover the matrix with corrupted entries.} algorithms for high-dimensional data processing, from simple to complex.

\subsubsection{Low-rank matrix recovery with dense noise and sparse anomalies}

Suppose we are given a large sensing data matrix ${\bf{Y}}$, and know that it may be decomposed as
\begin{align}
{\bf{Y}} = {\bf{X}}+{\bf{V}},
\end{align}
where ${\bf{X}}$ has low-rank, and ${\bf{V}}$ is a perturbation/noise matrix with entry-wise non-zeros. We do not know the low-dimensional column or row space of ${\bf{X}}$, not even their dimensions. To stably recover the matrix ${\bf{X}}$ from the sensing data matrix ${\bf{Y}}$, the problem of interest can be formulated as classical principal component analysis (PCA)~\cite{RPCA}:
\begin{align}
\label{eq:Opt1}
\min_{\{{\bf{X}}\}}~||{\bf{X}}||_*~~~{\text{subject~to}}~||{\bf{Y}}-{\bf{X}}||_F \le \varepsilon,
\end{align}
where $\varepsilon$ is a noise related parameter, $||\cdot||_*$ and $||\cdot||_F$ stands for the nuclear norm (i.e., the sum of the singular values) and the Frobenious norm of a matrix.

Furthermore, if there are also some abnormal data ${\bf{A}}$ injected into the sensing data matrix ${\bf{Y}}$, we have
\begin{align}
{\bf{Y}} = {\bf{X}}+{\bf{V}}+{\bf{A}},
\end{align}
where ${\bf{A}}$ has sparse non-zero entries, which can be of arbitrary magnitude. In this case, we do not know the low-dimensional column and row space of ${\bf{X}}$, not know the locations of the nonzero entries of ${\bf{A}}$, and not even know how many there are. To accurately and efficiently recover the low-rank data matrix ${\bf{X}}$ and sparse component ${\bf{A}}$, the problem of interest can be formulated as the following tractable convex optimization~\cite{Tracking-Network}:
\begin{align}
\label{eq:opt2}
&\min_{\{{\bf{X}},{\bf{A}}\}}~||{\bf{X}}||_{*}+\lambda ||{\bf{A}}||_{1}\nonumber\\
~~&{\text{subject~to}}~||{\bf{Y}}-{\bf{X}}-{\bf{A}}||_F \le \varepsilon,
\end{align}
where $\lambda$ is a positive rank-sparsity controlling parameter, and $||\cdot||_1$ stands for the $l_1$-norm (i.e., the number of nonzero entries) of a matrix.

\subsubsection{Joint matrix completion and matrix recovery}
In practical CIoT applications, it is typically difficult to acquire all entries of the sensing data matrix {\bf{Y}}, mainly due to i) transmission loss of the sensing data from the sensors to the data center, and ii) lack of incentives for the crowdsourcers to contribute all their sensing data.

In this case, the sensing data matrix $\widetilde{\bf{Y}}$ is made up of \emph{noisy}, \emph{corrupted}, and \emph{incomplete} observations,
\begin{align}
\widetilde{\bf{Y}}:={\mathcal{P}}_{\Omega}({\bf{Y}})={\mathcal{P}}_{\Omega}({\bf{X}}+{\bf{A}}+{\bf{V}}),
\end{align}
where $\Omega \subseteq [M]\times[N]$ is the set of indices of the acquired entries, and $\mathcal{P}_\Omega$ is the orthogonal projection onto the linear subspace of matrices supported on $\Omega$, i.e., if $(m,n) \in  {\Omega}$, ${\mathcal{P}}_{\Omega}({\bf{Y}}) ={y}_{m,n}$; otherwise, ${\mathcal{P}}_{\Omega}({\bf{Y}}) =0$. To stably recover the low-rank and sparse components ${\bf{X}}$ and ${\bf{A}}$, the problem can be further formulated as~\cite{Low-rank_MC}
\begin{align}
\label{eq:opt-com1}
&\min_{\{{\bf{X}},{\bf{A}}\}}~||{\bf{X}}||_{*}+\lambda ||{\bf{A}}||_{1}\nonumber\\
~~&{\text{subject~to}}~||{\mathcal{P}}_{\Omega}({\bf{Y}})-{\mathcal{P}}_{\Omega}({\bf{X}}+{\bf{A}}+{\bf{V}})||_F \le \varepsilon.
\end{align}

The problems formulated in (\ref{eq:Opt1}), (\ref{eq:opt2}), and (\ref{eq:opt-com1}) show the fundamental tasks of the research on ``matrix completion and matrix recovery" for high-dimensional data processing, which is receiving growing attention ranging from mathematicians to engineers (see e.g.,~\cite{Tracking-Network,RPCA,Low-rank_MC,ALM}). To efficiently solve the problems in (\ref{eq:Opt1}), (\ref{eq:opt2}), and (\ref{eq:opt-com1}), existing algorithms mainly include augmented Lagrange multipliers (ALM) algorithm and accelerated proximal gradient (APG) algorithm, which have been explained in~\cite{ALM} in detail. Readers can tailor the theoretical results in~\cite{Tracking-Network,RPCA,Low-rank_MC,ALM} and references therein to their specific CIoT applications of interest.

\subsection{Parallel and Distributed Data Processing}

So far, all the data processing methods introduced above are in essence centralized and suitable to be implemented at a data center. However, in many practical CIoT applications, where the objects in the networks are organized in an \emph{ad hoc} or decentralized manner, centralized data processing will be inefficient or even impossible because of single-node failure, limited scalability, and huge exchange overhead, etc. Now, one natural question comes into being: Is there any way to disassemble massive data into groups of small data, and transfer centralized data processing into decentralized processing among locally interconnected agents, at the price of affordable performance loss?

In this subsection, we argue that alternating direction method of multipliers (ADMM)~\cite{ADMM,BOOK-DB} serves as a promising theoretical framework to accomplish parallel and distributed data processing. Suppose a very simple case with a CIoT consisting of $N$ interconnected smart objects. They have a common objective as follows
\begin{align}
\label{ADMM1}
\min_{\bf{x}} f({\bf{x}})=\sum_{i=1}^N f_i({\bf{x}}),
\end{align}
where ${\bf{x}}$ is an unknown global variable and $f_i$ refers to the term with respect to the $i$-th smart object. By introducing local variables $\{{\bf{x}}_i\in {\bf{R}}^n\}_{i=1}^N$ and a common global variable ${\bf{z}}$, the problem in (\ref{ADMM1}) can be rewritten as
\begin{align}
\label{ADMM2}
&\min_{\{{\bf{x}}_1,...,{\bf{x}}_N,{\bf{z}}\}} \sum_{i=1}^N f_i({\bf{x}}_i)\nonumber\\
~~&{\text{subject~to}}~{\bf{x}}_i={\bf{z}}, ~~i = 1,...,N.
\end{align}

This is called the global consensus problem, since the constraint is that all the local variables should agree, i.e., be equal. The augmented Lagrangian of problem (\ref{ADMM2}) can be further written as
\begin{align}
\label{ADMM3}
&L_{\mu} ({\bf{x}}_1,...,{\bf{x}}_N,{\bf{z}},{\bf{y}})\nonumber\\
 &= \sum_{i=1}^N \big(f_i({\bf{x}}_i)+ {\bf{y}}_i^T({\bf{x}}_i-{\bf{z}})+\frac{\mu}{2}||{\bf{x}}_i-{\bf{z}}||_F^2  \big).
\end{align}

The resulting ADMM algorithm directly from (\ref{ADMM3}) is the following:
\begin{align}
&{\bf{x}}_i^{k+1} := \text{argmin}_{{\bf{x}}_i} \big( f_i({\bf{x}}_i)+ {\bf{y}}_i^{kT}({\bf{x}}_i-{\bf{z}}^k)+\frac{\mu}{2}||{\bf{x}}_i-{\bf{z}}^k||_F^2\big)\\
&{\bf{z}}^{k+1} := \frac{1}{N} \sum_{i=1}^N \big( {\bf{x}}_i^{k+1} +1/\mu {\bf{y}}_i^{k} \big)\\
&{\bf{y}}_i^{k+1}:={\bf{y}}_i^{k}+\mu( {\bf{x}}_i^{k+1}- {\bf{z}}^{k+1}).
\end{align}

The first and last steps are carried out independently at each smart object, while the second step is performed at a fusion center. Actually, when the smart objects are multi-hop connected, the second step can be replaced by
\begin{align}
\label{ADMM4}
{\bf{z}}_i^{k+1} := \frac{1}{|\mathcal{N}_i|} \sum_{i\in\mathcal{N}_i} \big( {\bf{x}}_i^{k+1} +1/\mu {\bf{y}}_i^{k} \big),
\end{align}
where $\mathcal{N}_i$ denotes the one-hop neighbor set of the $i$-th object and $|\cdot|$ is the cardinality of a set. Eq. (\ref{ADMM4}) means that the second step can also be carried out at each smart object by fusing the local data from one-hop neighbors.

This is a very intuitive algorithm to show the basic principle of ADMM. ADMM serves as a good general-purpose tool for optimization problems arising in the analysis and processing of massive datasets in a parallel and distributed manner. Apart from the intuitive ADMM algorithm for global consensus problem, more advanced topics include (but not limited to)~\cite{ADMM,BOOK-DB}:
\begin{itemize}
      \item Developing ADMM algorithms for distributed large-scale model fitting, where each update in subproblems (14)-(16) reduces to a model fitting problem on a smaller dataset. These subproblems can be solved using any standard algorithm suitable for small to medium sized problems. In this sense, ADMM builds on existing algorithms for single machines, and so can be viewed as a modular coordination algorithm that coordinates a set of simpler algorithms to collaborate to solve much larger global problems together than they could on their own.
      \item Implementation of ADMM algorithms in a MapReduce framework, where each iteration of ADMM can easily be represented as a MapReduce task: The parallel local computations are performed by Maps, and the global aggregation is performed by a Reduce. MapReduce is a popular programming model for distributed processing for very large datasets~\cite{MapReduce}.
\end{itemize}

\section{Semantic Derivation and Knowledge Discovery in Cognitive Internet of Things}
With massive data analytics, tremendous perceived data about physical world, cyber world, and social world in CIoT are well processed into an organized manner. However, as CIoT envisions trillions of objects to be connected and function cooperatively, it is still not feasible to utilize these analyzed data for decision-making directly due to both complexity and inefficiency. As one can imagine, only if the objects within CIoT are able to understand correctly and reason properly can they behave appropriately. For instance, the signal lamp in future smart transportation system may be able to understand how many vehicles and passengers are waiting at the intersection, whether there is an ambulance among them, which directions they are heading, and how long have they been waiting. These kinds of information are taken into consideration by the lamp, so as to decide how to change the transportation signal would be the most effective and fairest option. Besides, this signal lamp may figure out a few patterns after serving couple of months, such as the average time for 20 adult passengers to pass is about 30 seconds, or a bus always runs faster than a truck. This knowledge can not only be utilized by the lamp in future decision, but also be exported to the ones in social world.

\begin{figure}[!t]
\centering
\includegraphics[width=\linewidth]{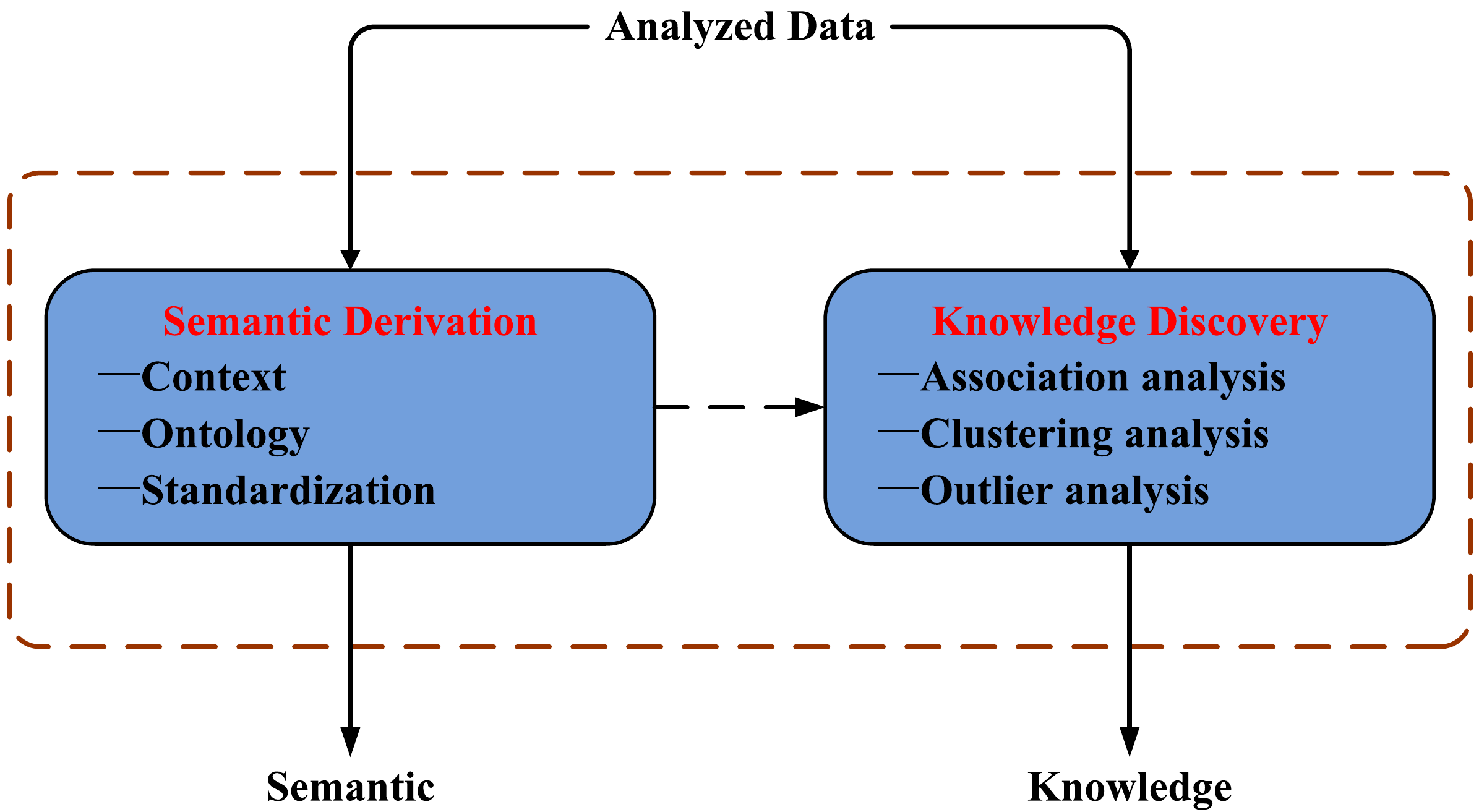}
\caption{The framework of semantic derivation and knowledge discovery.}
\label{Fengshuo}
\end{figure}

Therefore, to make the objects in CIoT understand and be aware, it is necessary to enable them to automatically derive the semantic from analyzed data. Besides, based on the analyzed data and semantic, some valuable patterns or rules can be discovered as knowledge as well, which is a necessity for everyday objects in CIoT to be, or appear to be intelligent, as illustrated in Fig.~\ref{Fengshuo}.

%%------------------------------------------------------------------------------------------
%%-----------------------------------------------------------------------------------------

\subsection{Semantic Derivation in CIoT}

Generally, semantic refers to the meaning of any (set of) object, situation, symbol, language, etc., and semantic derivation in CIoT is defined as the process of deriving semantic by adopting various kinds of semantic technologies from analyzed data. In this subsection, we introduce and discuss several key concepts in semantic derivation, i.e., context, ontology, and semantic standardization.

\subsubsection{Context in CIoT}

To date, there is no standard definition of context. A well known definition for context is provided by Abowd et al.~\cite{x1} as follows:

\emph{``Context is any information that can be used to characterize the situation of an entity. An entity is a person, place, or object that is considered relevant to the interaction between a user and an application, including the user and applications themselves."}

After massive data analytics, tremendous perceived data about physical world, cyber world, and social world are well organized. Once we put these analyzed data in such a way that they represent the situation of an entity, they are viewed as the context in CIoT. The context can be location, identity, time, activity, and so on. For example, the crowdsourced observations on the city road are perceived data. After analyzing, these data are organized and used to construct real-time traffic situation map and/or statistical traffic database, which are/is identified as context that characterizes the traffic situation of the city.

Although context in CIoT contains the semantic desired by the devices, it still needs to be further processed. One of the reasons is that in CIoT, the sources of context are massive and heterogeneous. As a result, one identical situation can be expressed in plenty of contexts from different sources, which in fact contains the same semantic. It promotes the difficulty of understanding the meanings for everyday devices. Since CIoT is envisioned to be capable of combining information of different contexts, it is necessary to apply effective semantic technologies to obtain the semantic from different contexts. Among others, ontology is treated as one of the most important component in semantic technologies, and will be discussed hereafter.

\subsubsection{Ontology in CIoT}
In the existing literatures (see, e.g.,~~\cite{Context-Survey} and \cite{x2}), it is established that one of the most appropriate formats to manage context is ontology. In philosophy, an ontology is a theory about the nature of existence, of what types of things exist; as a discipline ontology studies such theories. In artificial intelligence and Semantic Web researches, the term of ontology refers to a document or file that formally defines the relations among terms. In CIoT, the definition of ontology is adopted as Studer et al.~\cite{x4} have defined:

 \emph{``An ontology is a formal, explicit specification of a shared conceptualization. A conceptualization refers to an abstract model of some phenomenon in the world by having identified the relevant concepts of that phenomenon."}

Ontology offers an expressive language to represent the relationships and context, and has provided a solution to identify the same semantic came from different contexts in CIoT. For example, take identity as the context type in smart transportation, the contexts obtained by different objects might be: the highest buildings nearby, the most symbolic one, the one around corner, or the one with ATM machines in it. As a matter of fact, it is highly possible that all these contexts indicate the same semantic (meaning), which is the construction as headquarter of bank of communications. Note that by using various kinds of semantic technologies (whose extensive discussion falls out of the scope of this article) such as ontology, the semantic is derived from contexts comprised by analyzed data.

\subsubsection{Semantic Standardization in CIoT}

Undoubtedly, semantic standardization is an important enabler for the success of semantic derivation in CIoT paradigm, since it may effectively increase the semantic inter-operability and extendibility. CIoT supports interactions among massive heterogeneous sources of data and contexts through standard interfaces and models to ensure a high degree of semantic interoperability among diverse systems~\cite{roadmap}. Although many different semantic standards may coexist, the use of ontology based ones will enable mapping and cross-referencing between them, in order to enable information share/exchange.

In CIoT, semantic standards play an increasingly important role with every everyday objects connected. Its status is emphasized with two big changes occurred in CIoT: one is massive and heterogeneous sources in physical world, the other is tremendous and personalized application demands in social world. As a result, the objects are required to share and/or exchange semantic information continuously. Semantic standards make it possible for the objects to communicate the meanings with each other efficiently with minimum ambiguity.

Besides, semantic standardization can draw from, as well as serve for the intersected research field of CIoT and CRN. Standards regarding spectrum allocation, nodes selection, transmit power control, and communication protocols will ensure that the objects connected in CIoT/CRN can share the valuable radio spectrum with each other harmoniously. As greater reliance is placed on CIoT as the global infrastructure for processing information, it will be essential to deploy and further develop semantic standardization in future.

\subsection{Knowledge Discovery from Analyzed Data in CIoT}

As aforementioned in Section I-A, one of the major characteristics of CIoT compared to classic IoT is the emphasis on high-level intelligence. It is not accomplished in any separate part of CIoT, and should be considered throughout all the stages of design, development, implementation, and evaluation.

To achieve intelligence for the objects in CIoT, the most important way is to realize knowledge discovery from analyzed data, and then apply it in the following. In CIoT, knowledge is actually a broad concept that includes the general principles and natural laws related to every object. For example, the behavior (even thinking) patterns of human in social world, the correlations and functional mechanisms among all the components of cyber world, and the dynamic characteristics and common laws of physical world, and so on.

In general, knowledge is valid, certain, and potentially useful~\cite{x6}. It is also consolidated, contextualized, and more stable in time than data, context, or semantic in CIoT. As previous, take the smart transportation as an example. The crowdsourced observations on the street are regarded as raw data, the real-time traffic situation map and/or statistical traffic database are/is identified as analyzed data (context), the meaning of whether it is jammed on the way to destination currently is semantic, and the rules about that the average time for 20 adult passengers to pass is about 30 seconds, or a bus always runs faster than a truck are viewed as the knowledge in CIoT.

It is recognized that tremendous techniques from areas such as artificial intelligence, machine learning, pattern recognition, database technology, etc., can be applied to discover knowledge from analyzed data in CIoT. In this article, several knowledge discovery techniques which are well established in the above listed disciplines are introduced as follows~\cite{x7}.

\subsubsection{Association Analysis}

One of the feasible knowledge discovery techniques is association analysis. It is very useful for knowledge discovery from analyzed data, as there are many association types existing in CIoT.

\begin{itemize}
  \item \emph{Multilevel associations }involve semantic at different abstraction levels (such as the relation between street, city, and country understood by the objects in CIoT). To avoid achieving commonsense knowledge at high abstraction levels as well as avoid achieving trivial patterns at low or primitive abstraction levels, it is important to develop effective methods using multiple minimum support thresholds to discover this kind of association, with sufficient flexibility for easy traversal among different abstraction spaces.
  \item \emph{Multidimensional associations} involve more than one dimension (e.g., rules that relate how much time it costs for a passenger passing a street to width of the street and/or the passenger's age. Here, time, street width, and age are different dimensions). As the sources in CIoT are heterogeneous, the analyzed data obtained by different kinds of objects usually focus on different aspects. Techniques for analyzing such kind of associations should be chosen according to the difference in how they handle repetitive predicates.
  \item \emph{Quantitative association rules} involve numerical attributes or measures which have an implicit ordering among values (e.g., time, street width, or age as mentioned before). The quantitative attributes can be discretized into multiple intervals, and then be treated as nominal data in the discovery of this kind of association rules.
\end{itemize}

\subsubsection{Clustering Analysis}

Clustering analysis, as one of the classic knowledge discovery techniques, is the process of partitioning a set of analyzed data into subsets. Each subset is a cluster, such that the analyzed data in a cluster is similar to one another, yet dissimilar to the ones in other clusters. The set of clusters resulting from a cluster analysis can be referred to as a clustering. In CIoT, different clustering methods may generate different kinds of clusters on the same set of analyzed data. A simple example is that, the clusters formed when trying to count the number of vehicles and the number of passengers separately, are different from the clusters formed when the vehicles and passengers are partitioned by their heading directions. Since the partitioning is performed in cyber world rather than social world, clustering analysis is very useful to lead to the discovery of previously unknown groups within the analyzed data.

\subsubsection{Outlier Analysis}

Another useful knowledge discovery technique is outlier analysis. In CIoT, some of the objects may not comply with the general behavior or action model like others. These objects are considered as outliers. As the threats regarding security and privacy are extremely crucial in CIoT (such as falsifying the identity as an ambulance to pass the street early, or making credit card fraud to evade the payment for vehicles), outlier analysis plays an important role to keep CIoT from being compromised. To be specific, outliers may be detected using distance measures where objects that are remote from any cluster are considered as outliers, or using reputation mechanism where the reputation of one object is calculated based upon its neighbors' opinion or its historical behavior. After the outliers are differentiated from the normal ones, they would be discarded from CIoT to keep a pure environment.

Besides, CIoT also provides comprehensive reasoning mechanisms based upon ontology~\cite{x8}, which allows knowledge discovery from the derived semantic when it is necessary. Furthermore, the knowledge discovery process should be highly interactive. Thus, it is important to build flexible interfaces with social world, facilitating the user's interaction with the system in cyber world of CIoT. For instance, a user may like to first access to the derived semantic or discovered knowledge, evaluate its correctness and utility, and then modify or just regenerate it. Interactive discovery should allow the requests coming from social world to dynamically refine the discovery process, while taken the current situation of physical world into account. Knowledge that has been discovered as well as the input from social world should also be incorporated into the following knowledge discovery process as guidance. In addition, we should point out that the influence of knowledge is quite far-reaching~\cite{x9}, such as the influence of facilitation in data analytics, the influence of expectation in semantic derivation, and the influence of supplementation in decision-making, to name just a few.

\section{Intelligent Decision-making for Cognitive Internet of Things}
Generally, decision-making in CIoT includes reasoning, planing and selecting. For reasoning and planing, the key concerns are analyzing the collected data and inferring useful information, which belong to data analysis in essence. To avoid illegibility, we refer to decision-making as selecting in this article. The task of selecting is common in CIoT, e.g., selecting the path in the smart traffic systems, choosing the channels for wireless transmission, and selecting the optimal service when there are multiple services simultaneously available. To summarize, selecting can be defined as the process of choosing an action from the action set. Motivated by the learning ability in cognitive radio networks \cite{Haykin_2005}, we study \emph{cognitive selecting} in CIoT, which is characterized by having the ability to intelligently adjust the selecting based on the history information.

%%%%%%%%%%%%%%%%%%%%%%%%%%%
\begin{figure}[!b]
\centering
\includegraphics[width=\linewidth]{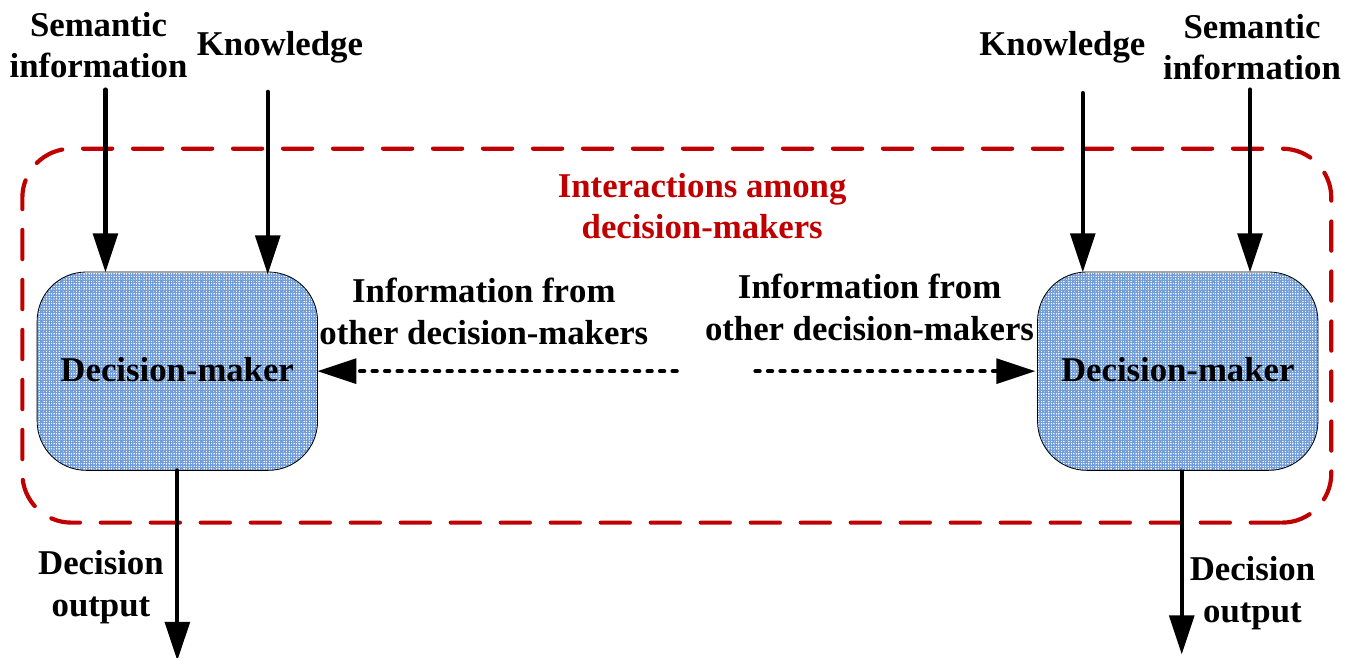}
\caption{The framework of intelligent decision-making in CIoT.}
\label{fig:framework}
\end{figure}

Methodically, three kinds of cognitive selecting have been  studied in the literature \cite{Xu_COMST_13}:  Markovian decision process, multi-bandit armed problem and multi-agent learning. In comparison, the first two kinds are mainly for single decision-maker while the third one is for multiple decision-makers. Since it is expected that there are a large number of decision-makers (human or machine) in CIoT, we focus on multi-agent learning. Since the selections of the decision-makers are interactive, we can formulate the multiple decision-making system as a game and then study multi-agent learning approaches. Specifically, we establish a framework for intelligent decision-making in CIoT, study intelligent decision-making in large-scale CIoT, and investigate the learning approaches with uncertain, dynamic and incomplete information.

\subsection{A Framework for Intelligent Decision-Making in CIoT}
We establish a framework for intelligent decision-making in CIoT, which is shown in Fig. \ref{fig:framework}. Each decision-maker has semantic information and/or knowledge from the environment. Note that semantic information is generally  generated by semantic derivation, while knowledge can be obtained from knowledge discovery or be given from social world in advance. In addition, it may have information about other decision-makers if information exchange is available. However, if the information exchange is resource-consumed or even not available in some scenarios, a decision-maker does not have information about others. Using the information from the environment and others and taking into account  its service demand, a decision-maker performs the cognitive selecting and outputs the decision result.

Since there are multiple decision-makers in CIoT, the selections of the decision-makers are interactive. To capture the interactions among multiple decision-makers, one would formulate the problems of cognitive selecting as game models, which were originally studied in economy and have been successfully applied into several engineering fields. To get a better understanding of game models, we briefly present the game model. Formally, a game is denoted as $\mathcal{G}=\{\mathcal{N},A_n,u_n\}$, where $\mathcal{N}$ is the set of players, $A_n$ is the selection set of player $n$ and $u_n$ is its utility function. Due to the feature of distributed and autonomous decision-making in CIoT, non-cooperative game models can characterize the interactions among decision-makers well. In a non-cooperative game, each player maximizes its individual utility function and Nash equilibria are the well-known stable solutions for non-cooperative games. A pure strategy Nash equilibrium is a selection profile such that if and only if no player can improve its utility function by deviating unilaterally. Other concepts of stable solutions in non-cooperative games are correlated equilibria \cite{Foster_GEB_97} and Bayesian Nash equilibria. For more analysis of game models, refer to \cite{game_book}.

Technically, there are two important issues for using game-theoretic learning for CIoT \cite{Young_GEB_09, Pradelski_GEB_13}:
\begin{itemize}
  \item Designing utility function. It is emphasized that a game model only addresses the interactions among multiple decision-makers, whereas it does not guarantee the performance. In some worse cases, the selfish nature of players may lead to inefficiency and dilemma, which is known as \emph{tragedy of commons} \cite{Kameda_JSAC_08}. Thus, one should carefully design the utility functions such that some metrics  can be improved, e.g., the aggregate quality of experience (QoE), fairness and resource utilizing efficiency. Moreover, the designed utility function should admit some stable solutions, which are important for practical applications.
  \item Achieving the stable solutions. With the multiple decision-making problem in CIoT now formulated as game models, learning procedures are needed to converge to the desirable stable solutions. In particular, appropriate approaches are desirable for solving different information constraints, e.g., the network may be static or dynamic, the system parameters may be known or unknown, the information about the environment and others may be complete or incomplete.
\end{itemize}

\subsection{Intelligent Decision-Making in Large-Scale CIoT}
\begin{figure}[!t]
\centering
\includegraphics[width=0.6\linewidth]{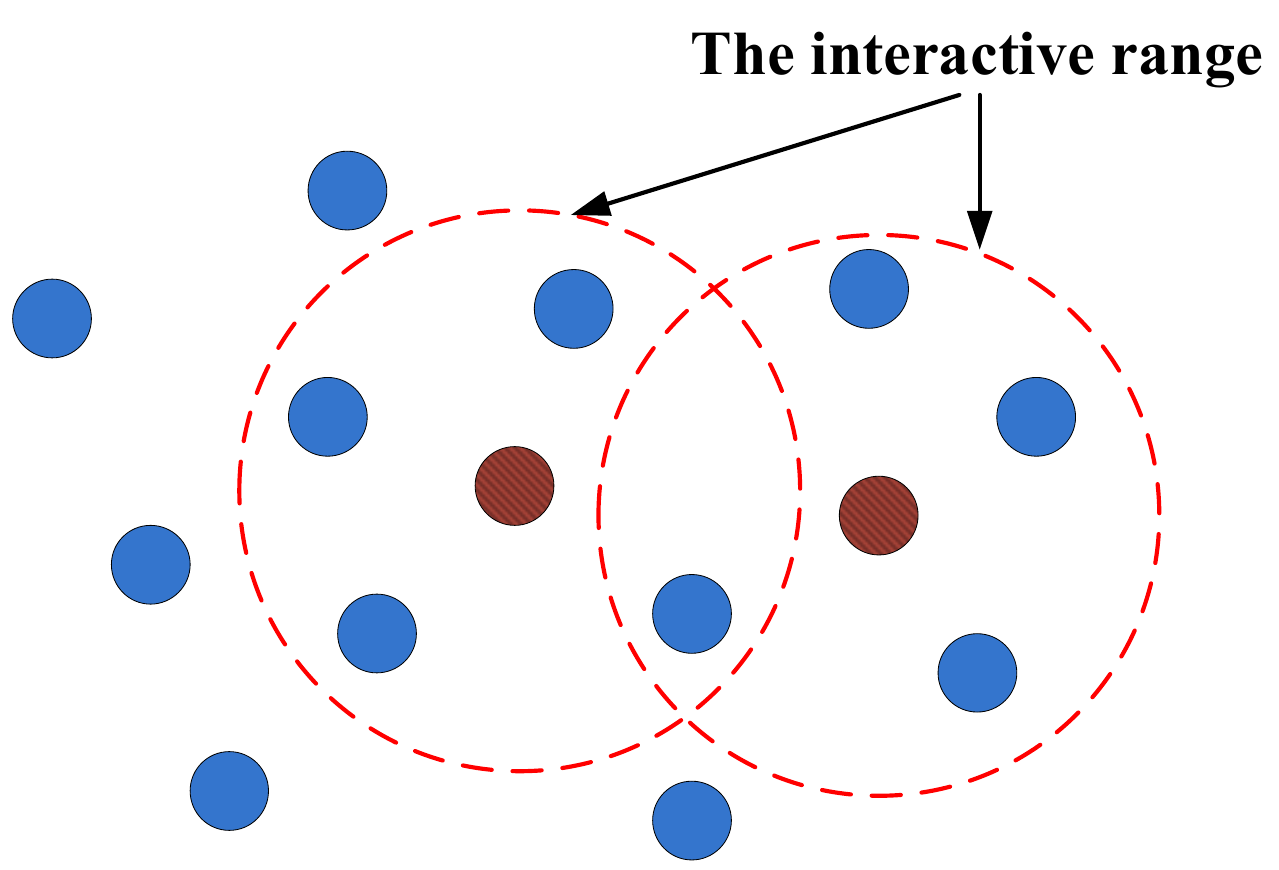}
\caption{An illustrative example of local interactions in CIoT.}
\label{fig:local_interaction}
\end{figure}

An interesting feature of CIoT is that there are always large number of spatially distributed decision-makers. Moreover, the selection of a decision-maker only has direct impact on its nearby decision-makers; that is, the decisions in CIoT are local interactive.
Examples of local interactions are given by: the vehicles in the smart traffic systems only affecting other vehicles in proximity and
a sensor contending for resources (channel, energy and time) only with its neighbors. An illustrative example of local interactions in CIoT is shown in Fig. \ref{fig:local_interaction}. The interactive range is context-dependent. There may be overlapping in the interactive ranges, which is determined by the network topology; moreover, when all other decision-makers are located in the interactive range of each decision-maker, the local interaction becomes ordinary global interaction.

The feature of local interaction makes the corresponding game models different. Specifically, the game is called spatial game, which is denoted as $\mathcal{G}_l=\{\mathcal{N}, \mathcal{J}_n, A_n, u_n\}$, where $\mathcal{N}$ is the set of players, $\mathcal{J}_n$ is the player set in the interactive range of $n$, $A_n$ is the selection set of player $n$ and $u_n$ is its utility function. In global interactive games, the utility function is determined by the selection profiles of all players, i.e., the utility function is expressed as $u_n(a_n,a_{-n})$, where $a_{-n}$ is the selection profiles of all other players except $n$. In comparison, the utility function in spatial game is expressed as
$u_n(a_n,a_{\mathcal{J}_n})$, where $a_{\mathcal{J}_n}$ is the selection profiles of players in the interactive range of $n$.

It is seen that spatial game models are more suitable for large-scale CIoT. However, due to the spatially distribution of players, spatial game is generally hard to analyze. The main reasons are: 1) although the direct interactions are local, there  exists inherent mutual interaction among any two arbitrary players, 2) the number of all players may be huge, 3) the interactive neighbors of players are different. Thus, to make the game-theoretic approaches feasible in large-scale CIoT, some efforts are needed. One promising approach is introducing local cooperation into spatial games. Specifically, although global information exchange among all players is not possible in large-scale CIoT, local information in the interactive range is feasible. Based on this, the player behaves altruistically by taking its interactive neighbors into account. It was shown in \cite{Xu_JSTSP_12} that local cooperation leads to near-optimal optimization.

\subsection{Decision-Making with Uncertain, Dynamic and Incomplete Information}
In this subsection, we investigate information constraints in CIoT, which are important for decision-making problems. Specifically, we study intelligent decision-making with \emph{uncertain}, \emph{dynamic} and \emph{incomplete} information. The presented three information constraints are common in CIoT. Taking the smart traffic systems as an example, the arrival of vehicles is random, the congestion level of a path is dynamic and a sensor may have partial  information for an event and have no information about others sensors.

To deal with the above information constraints, the optimization metrics should be carefully designed. Generally, there are two optimization metrics with uncertain, dynamic and incomplete information. The first is to maximize the expected payoff, i.e. $\max \text{E}[r_n(t)]$, where $r_n(t)$ is the random payoff after each play. The second is to minimize the outage probability, i.e., $\min \Pr \{r_n(t)>\eta_n\}$, where $\eta_n$ is the threshold for achieving certain service. To illustrate, one may want to minimize the expected traveling time from  home to the office, or minimize the probability that the traveling time is large than thirty minutes.

\begin{figure}[!t]
\centering
\includegraphics[width=0.9\linewidth]{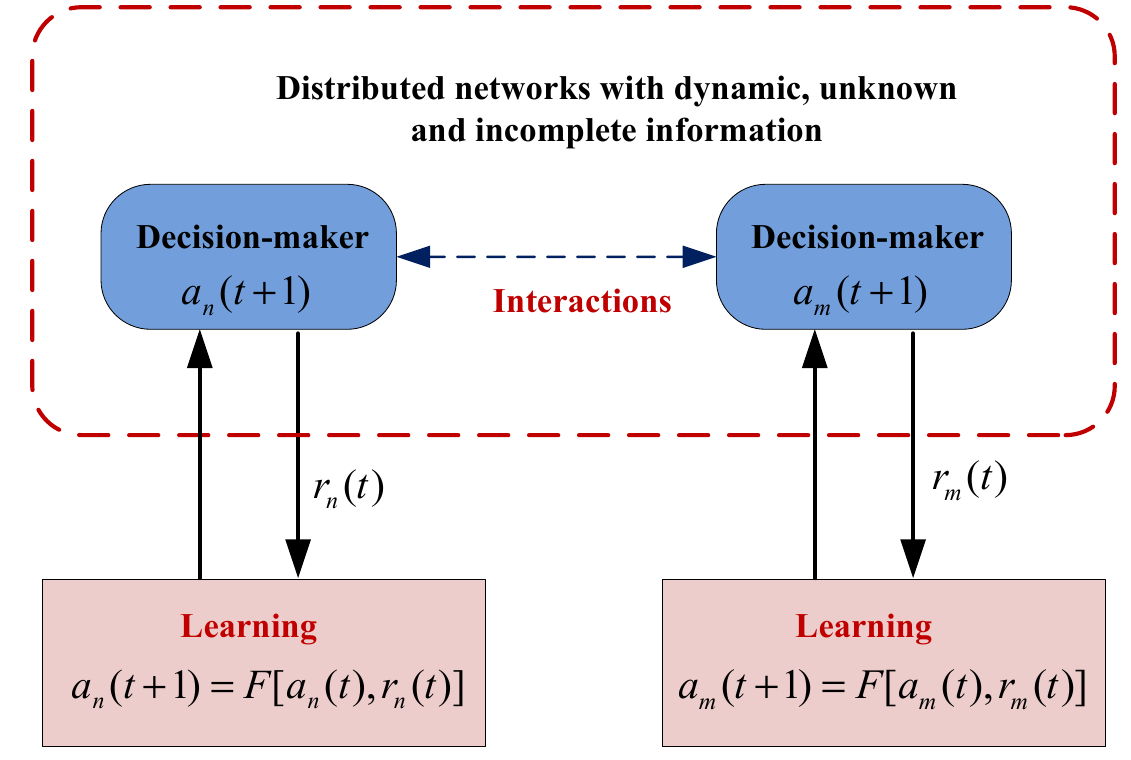}
\caption{The illustrative diagram of learning with uncertain, dynamic and incomplete information in CIoT.}
\label{fig:learning}
\end{figure}

To solve the  uncertain, dynamic and incomplete information constraints, learning is a promising approach. The illustrative diagram of learning  is shown in Fig. \ref{fig:learning}. The ideas are as follows: 1) for a given selection profile $(a_n(t),a_{-n}(t))$, each player gets a random payoff $r_n(t)$, which is jointly determined by the selection profile and the environment,
2) the players employ a rule to update its next selection based on the current selection and payoff, i.e., $a_n(t+1)=F[a_n(t),r_n(t)]$,
3) this procedure is repeated until some stopping criterion is met.  It is noted that the proposed learning scheme is autonomous and fully distributed, since it only relies on the individual history information of a player; moreover, its convergence property can be analyzed by the theories of Markovian process and stochastic approximation \cite{Sastry_SMC_94}.

%For different applications in large-scale CIoT, the game models and the multi-agent learning algorithms should be carefully designed. In particular, the local interaction and the uncertain, dynamic and incomplete information constraints should be taken into account for decision-making.

\begin{figure}[!t]
  % Requires \usepackage{graphicx}
  \centering
  \includegraphics[width=0.8\linewidth]{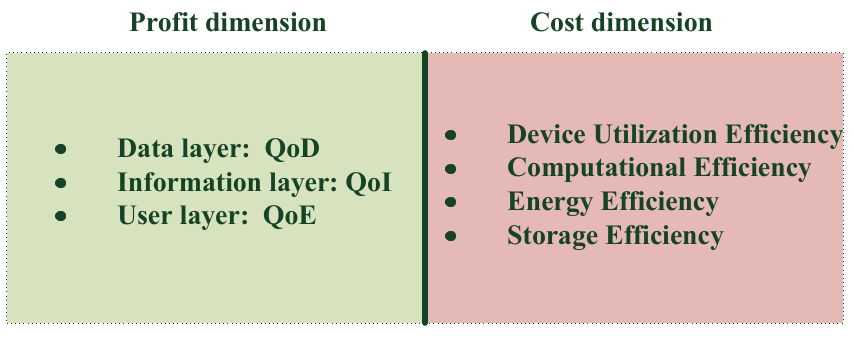}\\
  \caption{Considered metric structure.}\label{fig:metric}
\end{figure}

\section{Performance Evaluation Metrics in Cognitive Internet of Things}

Evaluating the performance of CIoT service is a challenging task, since a lot of considerations and factors are involved. In order to fully cover the issue, we broadly divide the metrics into two dimensions: profit and cost. The profit dimension corresponds to appealing results in CIoT, while the cost dimension considers the cost efficiency aspect. The overall structure of considered metrics is presented in Fig. \ref{fig:metric}.

\subsection{Profit dimension}

As shown in Fig. \ref{fig:metric}, we expect to characterize the profit dimension from the following three layers: data layer, information layer and user layer. Corresponding to these three layers, we use three metrics quality of data (QoD), quality of information (QoI) and quality of experience (QoE), respectively.

\subsubsection{Data layer-QoD}

The data layer metric aims to evaluate the quality of sensed data, the process of data acquiring and the possible data distribution at the Perception/Sensing stage. Data plays a fundamental role in the CIoT cycle and evaluating its quality is desirable.

In IoT, the acquired data may not meet system requirement resulted from the following factors. Firstly, the data is commonly noisy in practice, due to the environment noise and sensing devices' deviation and limited sensing accuracy. The sensing accuracy and deviation may vary from device to device. Secondly, the data may be corrupted by malicious data. Thirdly, the data can be incomplete since not all data can be collected considering the limited number of sensor devices and constrained sensing cost. Furthermore, even the data is accurate and complete, it can be outdated for demand. In response to above considerations, we propose a new metric, quality of data (QoD). The QoD consists of data accuracy, data truthfulness, data completeness and data up-to-dateness \cite{QoC2}. Apparently, the data accuracy reflects the precision of collected data. The data truthfulness indicates the reliability degree of the data resource. The data completeness corresponds to the ratio of collected data amount to the amount of all required data. The data up-to-dateness reflects the validity of data to the decision making, i.e., if the data is too late to assist decision-making, it is meaningless. The four aspects jointly determine the overall quality of data.

%---------------------------------------------------------------------------------------------------------------------------------------------------
\begin{figure*}[!t]
\centerline{
\subfloat[$\text{QoE framework in IoT}$.]{\includegraphics[width=2in]{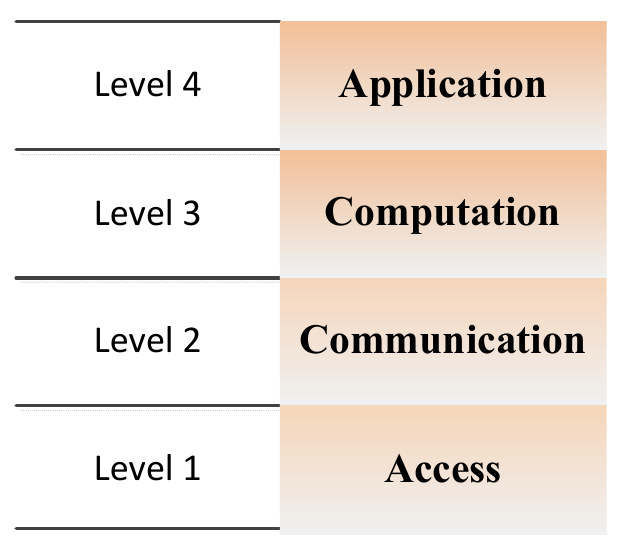}
\label{fig:QoE}}
\hfil
\subfloat[$\text{Different service provisioning level-QoE mappings}$.]{\includegraphics[width=2.5in]{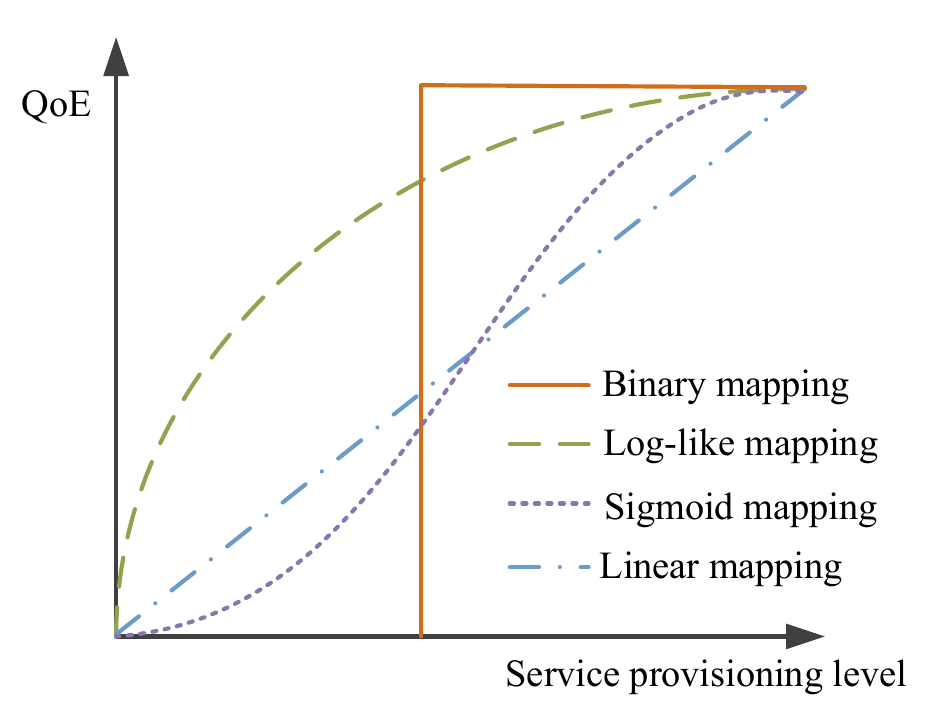}
\label{fig:mapping}}
}
\caption{QoE in user layer.}
\label{fig-QoE0}
\end{figure*}
%---------------------------------------------------------------------------------------------------------------------------------------------------

\subsubsection{Information layer-QoI}

Since CIoT is marked with the intelligent decision-making, where information plays a key role in functional Cycle of CIoT, the quality of information in decision-making needs to be evaluated. We treat the information as the input to decision making and resort to the concept of quality of information (QoI) in \cite{4}. We believe that QoI is a satisfactory metric at present, as it tries to concern the information that meets decision maker's need at some place, location, social setting and specific time. Existing QoI metric is defined as
\begin{align}
QoI=Q*P*R*A*D*T*V,
\end{align}
where $Q$ denotes quantity, $P$ denotes precision, $R$ denotes recall, $A$ denotes accuracy, $D$ denotes detail, $T$ denotes timeliness, $V$ denotes validity. All values are normalized into $\left[ 0, 1 \right]$ with 1 representing the corresponding best case.

In the above metric, quantity represents how much useful information the decision maker has obtained for a specific task. If all needed information is available, $Q=1$. Precision here may refer to the proportion of relevant information to all information gathered by sensors, networks or services. On the other hand, recall refers to the proportion of relevant information without the assistant from sensors, networks or services. Accuracy represents the accuracy degree of information to decision maker's requirement. Note that quantity, precision, recall and accuracy jointly characterize the quality of the information quantity provided. Detail characterizes the complete degree of the information to the decision maker. Timeliness is used to measure the decision maker's timeline along which the information is to be employed. We denote the time delay as the gap between the instant the information available and the instant the information employed. Then, the timeliness can be treated as inversely proportional to the time delay. If the information is available before the decision-maker using it, the timeliness is 1. Validity reflects the trueness of the provided information. We may find that the QoD and QoI share some similar properties. However, the involved objectives are different, that is, QoD is used for data quality evaluation, while the QoI is used for information quality evaluation.

\subsubsection{User layer-QoE}

QoE is defined by the International Telecommunication Union (ITU) as ``the overall acceptability of an application or service, as perceived subjectively by the end user'' \cite{2}. Since IoT mainly concerns applications for human, we believe the ideal of QoE is suitable for measuring user profit in IoT applications. While existing QoE in communications and networking is mostly derived from communication quality provisioning \cite{3}, neglecting the role of upper layer computation resource and application quality. Therefore, we extent the QoE concept and derive a new QoE framework as shown in Fig. \ref{fig-QoE0}(a).

In the proposed framework, QoE is evaluated from factors in four levels. Specially, level 1 ``Access'' focuses on the basic Internet connection ability of application related things and objects, since without the Internet connection, the IoT almost losses its spirit. Upon the access ability, level 2 turns to the communication capability to guarantee the running of application. Clearly, different applications or traffic may impose diverse communication capabilities. Both level 1 and level 2 capture the impact of communication on QoE, which generally corresponds to existing QoE modeling methods. On the other hand, level 3 turns the focus to computation ability, which brings the computation resource, the new emerging resource from cloud computing, into consideration. Note that this is important for computation-intensive applications in CIoT. Finally, applications directly deliver the service to human. For example, whether the user interface is friendly and whether the service is custom for human can greatly affect human's perception. Thus, ``application'' constitutes level 4.

Define the overall performance of the above five levels as service provisioning, different service provisioning level to QoE mappings can reflect users' heterogeneous demands to some extent. As shown in Fig. \ref{fig-QoE0}(b), the different curves indicates users' diverse elasticity \cite{inelastic1}\cite{inelastic2}  or sensibility to service provisioning.

\subsection{Cost dimension}

There is ``no free lunch'' in that for every gain we make in practice there is a price to be paid. Hence, we also consider the cost dimension metric in terms of resource efficiency. In particular, the resource efficiency embodies four types as follows.
\begin{itemize}
  \item \emph{Device Utilization Efficiency:}
  Hardware resource, especially device resource is commonly constrained for two reasons. On one hand, some devices, for example, the spectrum analyzer, are expensive. On the other hand, even for some relatively cheap sensors, ideally dense deployment is unrealistic. Therefore, given limited device resource, maximizing the utilization or exploring the capability of devices is indispensable. The device utilization efficiency evaluates the degree of utilization efficiency given limited devices resource. For example, given the same amount of carbon dioxide sensors, increasing the sensors' sample points on the geographic area by moving sensors can increase the gathered data amount, compared with static sensors deployment cases. Correspondingly, the sensors's utilization efficiency is improved in the moving sensors case.
  \item \emph{Computational Efficiency:}
  Computation and decision making in CIoT incur computational load. The computing resource will become scarce when a large number of users with various computation intensive tasks are involved in CIoT. Although the cloud computing paradigm can relieve the computing resource scarce problem, improving the computational efficiency is still a fundamental requirement.
  \item \emph{Energy Efficiency:}
   The energy consumption in CIoT may occur at all cognitive tasks from perception to decision-making, action/adaptation and service provisioning. Note that besides the conventional energy consumption considerations in communication systems such as cellular network, additional energy is needed to support massive and ubiquitous wireless access for connected things, considerable computation and storage ability. In CIoT, the number of connected things on the global will be numerous, resulting in a considerable consumption in energy. Therefore, there is a urgent demand to improve the energy efficiency in IoT. The energy efficiency metric has to be able to reflect overall energy utilization level incorporating communication, computation, storage, etc., in order to provide quantified information to reflect energy consumption of certain configurations, to compare energy consumption performance of different applications and solutions, to set research and development targets on energy efficiency \cite{5}.
  \item \emph{Storage Efficiency:}
  Storage cost is another important aspect of cost, as the storage and update of data, information and knowledge rely on physical storage in CIoT. With the increasing larger amount of information and data in the emerging big data era, the storage demand will increase in CIoT, and the storage problem is becoming a new challenge. Storing the largest amount of data with the least physical storage and without performance loss to the service is always preferred. Thus, the storage efficiency evaluates the ability to store and manage data given fixed amount of physical storage space.
\end{itemize}

\section{Research Challenges and Open Issues}
CIoT has truly drawn a beautiful and exciting future, though current researches and developments are still far away from that vision. Several major research challenges and open issues include (but not limited to):
\begin{itemize}
  \item In practical CIoT applications, it is much more challenging to process the obtained massive sensing data that can be of \emph{mixed characteristics}, including heterogeneity, high-dimensionality, and nonlinear separability, etc.
  \item For different applications in large-scale CIoT applications, the game \emph{models} and the multi-agent learning \emph{algorithms} should be carefully designed. In particular, the local interaction and the uncertain, dynamic and incomplete information constraints should be taking into account for decision-making.
  \item In most existing multi-agent learning algorithms, the players update their strategies based on the history action-payoff information. This procedure may take long time to converge since the players need to explore all the possible selections. In CIoT, some new \emph{knowledge-assisted learning technologies} should be developed to increase the converging speed and achieve better performance.
  \item Developing effective semantic technologies and knowledge discovery techniques that are more suitable for CIoT applications is still a fundamental task.
  \item Most of current studies on QoE are limited on single user case, there is lack of study on system-level QoE, especially for large scale CIoT systems with massive users.
  \item Generic approaches in CIoT research mainly focus on abstracting \emph{common techniques} involved in various applications. However, generic approaches cannot be directly used for each specific situation. To apply the generic approaches for \emph{specific situations}, more practical constraints should be further considered.
  \item Last but not least, more attention should be focused on building the bridge from theory to practice. For example, how and where might the theoretical studies in CIoT research actually be applied? What does it mean for the company implementing a smart city?
\end{itemize}

\section{Concluding Remarks}
A new network paradigm, named Cognitive Internet of Things (CIoT), was developed in this paper to empower the current IoT with a `brain' for high-level intelligence, where general objects can not only see, hear, and smell the physical world for themselves, but also learn, think, and understand physical and social worlds by themselves. Inspired by human cognition process, we first presented a comprehensive definition for CIoT. Based on this definition, we further provided an operational framework of CIoT, which characterizes the fundamental cognitive tasks. Then, we addressed the key enabling techniques involved in the cognitive tasks in detail. In addition, we also discussed the design of proper performance evaluation metrics and the research challenges and open issues ahead.

Finally, we envision that the presented research is offered as a mere baby step in a potentially fruitful research direction. We hope that this article, with interdisciplinary perspectives, will stimulate more interests in research and development of CIoT, to enable smart resource allocation, automatic network operation, and intelligent service provisioning.

% use section* for acknowledgement
\section*{Acknowledgment}
We thank the editor and anonymous reviewers for their constructive comments and suggestions, which help us improve the presentation of this paper. We also thank Prof. Yueming Cai, Mr. Ducheng Wu, Long Wang, Junfei Qiu, Liang Yue, Zhen Xue, and Ms. Yijie Luo for their helpful discussions.

\end{document}